\documentclass[journal,twoside,web]{ieeecolor}
\usepackage{jsen}
\usepackage{amsmath,amssymb,amsfonts}
\usepackage{algorithmic}
\usepackage{graphicx}
\usepackage{wrapfig}
\usepackage{multirow}
\usepackage{color, colortbl}
\definecolor{LightCyan}{rgb}{0.88,1,1}
\definecolor{LimeGreen}{rgb}{0.47,0.71,0.04}
\usepackage{subcaption}
\usepackage{xcolor}
\usepackage[colorlinks=true]{hyperref}

\newcommand\scalemath[2]{\scalebox{#1}{\mbox{\ensuremath{\displaystyle #2}}}}
\def\BibTeX{{\rm B\kern-.05em{\sc i\kern-.025em b}\kern-.08em
    T\kern-.1667em\lower.7ex\hbox{E}\kern-.125emX}}
\definecolor{abstractbg}{rgb}{0.89804,0.94510,0.83137}
\begin{document}
\title{Joint Learning for Scattered Point Cloud Understanding with Hierarchical Self-Distillation}
\author{Kaiyue Zhou, \IEEEmembership{Member, IEEE}, Ming Dong \IEEEmembership{Member, IEEE}, Peiyuan Zhi, and Shengjin Wang, \IEEEmembership{Senior Member, IEEE}
\thanks{This work was supported by the National Key Research and Development Program of China in the 14th Five-Year (Nos. 2021YFF0602103 and 2021YFF0602102). (Corresponding author: Shengjin Wang.)}
\thanks{Kaiyue Zhou, Peiyuan Zhi, and Shengjin Wang are with the Department of Electronic Engineering, Tsinghua University, Beijing 100084, China (email: \href{mailto:kyzhou@wayne.edu}{kyzhou@wayne.edu}; \href{mailto:zhipy20@mails.tsinghua.edu.cn}{zhipy20@mails.tsinghua.edu.cn}; \href{mailto:wgsgj@tsinghua.edu.cn}{wgsgj@tsinghua.edu.cn}).}
\thanks{Ming Dong is with the Department of Computer Science, Wayne State University, Detroit, MI 48202, USA (email: \href{mailto:mdong@wayne.edu}{mdong@wayne.edu}).}
}

\IEEEtitleabstractindextext{%
\fcolorbox{abstractbg}{abstractbg}{%
\begin{minipage}{\textwidth}%
\begin{wrapfigure}[12]{r}{3in}%
\includegraphics[width=3in]{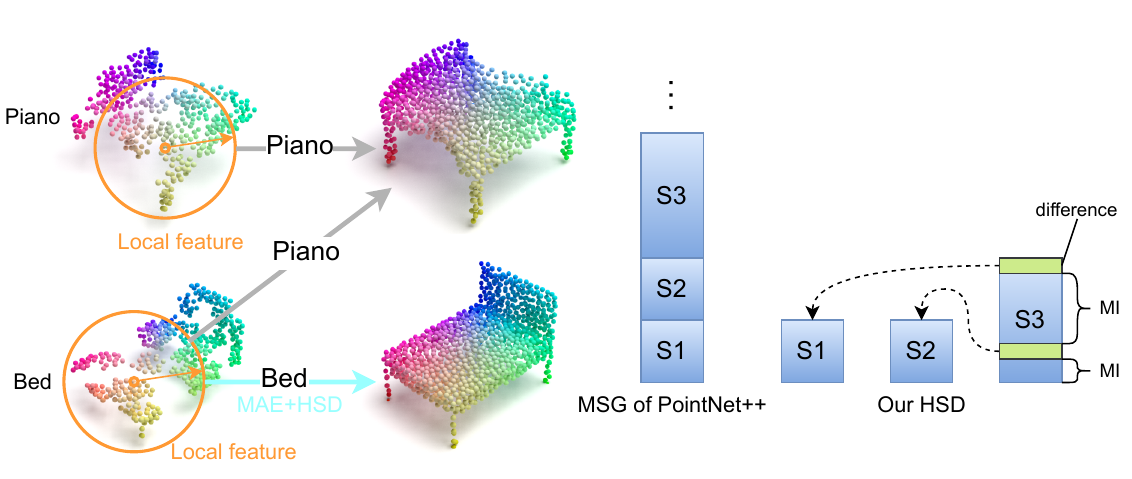}%
\end{wrapfigure}%
\begin{abstract}
Numerous point-cloud understanding techniques focus on whole entities and have succeeded in obtaining satisfactory results and limited sparsity tolerance. However, these methods are generally sensitive to incomplete point clouds that are scanned with flaws or large gaps. In this paper, we propose an end-to-end architecture that compensates for and identifies partial point clouds on the fly. First, we propose a cascaded solution that integrates both the upstream masked autoencoder (MAE) and downstream understanding networks simultaneously, allowing the task-oriented downstream to identify the points generated by the completion-oriented upstream. These two streams complement each other, resulting in improved performance for both completion and downstream-dependent tasks. Second, to explicitly understand the predicted points' pattern, we introduce hierarchical self-distillation (HSD), which can be applied to any hierarchy-based point cloud methods. HSD ensures that the deepest classifier with a larger perceptual field of local kernels and longer code length provides additional regularization to intermediate ones rather than simply aggregating the multi-scale features, and therefore maximizing the mutual information (MI) between a teacher and students. The proposed HSD strategy is particularly well-suited for tasks involving scattered point clouds, wherein a singular prediction may yield imprecise outcomes due to the inherently irregular and sparse nature of the geometric shape being reconstructed. We show the advantage of the self-distillation process in the hyperspaces based on the information bottleneck principle. Our method achieves state-of-the-art on both classification and part segmentation tasks. 
\end{abstract}

\end{minipage}}}

\maketitle

\section{Introduction}

Point cloud completion and understanding are essential in quite a number of 3D computer vision tasks, including autonomous driving, augmented reality, industrial manufacturing, and robotic maneuver. Especially, in interactive scenarios such as robotic grasp, an ill-scanned object or scene leads to difficulty on accurate interactions for two reasons. First, planes and objects need to be reconstructed from sensor-produced point clouds for collision detection and grasp pose selection. Second, the interaction manner and force are dependant upon the cognition of objects that is subject to the irregularity and sparseness of point clouds. 
In recent years, numerous studies~\cite{qi2017pointnet, qi2017pointnet2, wang2019dynamic} were proposed to leverage point cloud directly for understanding. 
While the globally- and locally- extracted features increase the tolerance of irregularity and sparseness, they are hindered from processing point clouds with a large amount of missing parts or overly sparse surfaces due to fast scanning, occlusion, view point, or sensor resolution. As a result, \cite{yan2021sparse, jaritz2019multi} were proposed specifically concentrating on sparse point clouds to tackle this problem. However, these methods utilizing either sophisticated voxelization procedure or multi-view photometric information do not generalize well due to limited memory or the need of additional supervisory signal.

An intuitive plan is to first compensate the incomplete parts, and then to sequentially apply a recognition algorithm on the completed point clouds. Nevertheless, models trained separately may suffer from lower robustness caused by the inconsistency in point distributions. Without explicitly learning the overall shape, the network may allocate excessive attention to irrelevant points or outliers caused by irregular local shapes, leading to degraded performance. 
The masked autoencoder (MAE) was initially developed to pre-train encoders by learning the correspondence between masked sections and original input data such as texts, images, or point clouds. This facilitated fine-tuning processes for natural language processing (NLP)~\cite{devlin2019bert} and computer vision~\cite{he2022masked, xie2022simmim, bao2022beit} tasks. Albeit MAE is primarily regarded as a pre-training model for improving classification performance, its reconstruction capacity is advantageous in scenarios where observations are incomplete. 
Additionally, hierarchy-based networks are designed to aggregate sparse features learned from multiple scales for better coverage of local regions. 
However, the aggregation operation may result in the aggregated representation being unaware of the mutual information across scales. As a solution, Deeply-Supervised Nets (DSN)~\cite{lee2015deeply} provides separate guidance to each scale as a strong regularization technique, but is still unaware of the correlation between those scales. 

In order to tackle the challenges inherent in scattered point cloud classification, we propose an approach based on the previous work ~\cite{zhou2024cascaded} for jointly learning object completeness and shape category through hierarchical self-distillation (HSD). In addition, we extend this approach to part segmentation by developing an adapted version of HSD that enhances point-wise understanding using richer shape representations. Our network consists of two cascaded streams: upstream and downstream. In the upstream, we generate portions of a point cloud by selecting random centroids as unmasked parts, after which a MAE learns the asymmetric mapping from unmasked (complete) to masked (incomplete) point clouds. In the downstream, we can use any hierarchy-based point cloud classifier, such as PointNet++~\cite{qi2017pointnet2}, CurveNet~\cite{xiang2021walk}, or PointMLP~\cite{ma2022rethinking}. To further regularize training, our HSD framework maximizes mutual information across multiple coverage scales and transfers the most distinctive features from the deepest hyperspace to other scales. Specifically, the classifier at the last scale serves as the teacher to supervise preceding classifiers, thus transferring critical knowledge across different hyperspaces. We demonstrate that DSN can be viewed as a special case of HSD and leverage the information bottleneck principle to explain our model, validating its generalization performance through experiments.
Our method emphasizes joint learning, where both MAE and HSD harmonize distributions to improve reconstruction and understanding. Our primary \textbf{contributions} are summarized as follows: 
\begin{itemize}
\item An end-to-end partial point cloud processing network is proposed to jointly learn the geometric features from irregular shapes, which dedicates to leverage the reconstruction with low-level shapes to help understanding, and in turn, to use understanding of high-level details to help the reconstruction.
\item We introduce a hierarchical self-distillation (HSD) framework that maximizes the mutual information captured in scale-aware hyperspaces. With its plug-and-play flexibility, HSD can be integrated into any hierarchy-based point cloud baseline.
\item Our evaluations on scattered point clouds show that our method performs competitively on ModelNet40, achieving state-of-the-art (SOTA) results on the real-world ScanObjectNN dataset for classification and on ShapeNetPart for part segmentation.
\end{itemize}


\section{Related Work}


\subsection{Masked Autoencoder. }
Initially, the masked autoencoder (MAE) was introduced as a self-supervised solution to handle the lack of accessible labels in data for NLP~\cite{devlin2019bert}. Although the information density in computer vision is different from that of NLP, researchers have shown that MAE works effectively in computer vision by masking out certain portions of pixels in the asymmetric architecture to effectively learn largely missing patches in the pixel space~\cite{he2022masked, xie2022simmim, bao2022beit}. Following the image works, PointMAE~\cite{pang2022masked} was the first MAE method for point clouds. Based on PointMAE, more 3D related MAEs~\cite{liu2023inter, jiang2023masked} have been proposed.

These methods, however, treat MAE as a pre-training model, requiring a fine-tuning stage for new data prediction. Our aim is to tackle incomplete object classification in an end-to-end paradigm, requiring only the process of self-supervised reconstruction in MAE. The completion networks~\cite{yuan2018pcn, tchapmi2019topnet, wen2020point, wen2021pmp, xiang2021snowflakenet, wen2022pmp, zhao2022pcunet} are well-suited for this purpose, as the synthetically pre-processed incomplete data can be viewed as naturally masked point clouds. SnowflakeNet~\cite{xiang2021snowflakenet}, operating on point splitting, achieves satisfactory reconstruction results with a relatively simple structure. In this study, we adopt the simplified SnowflakeNet as a masked autoencoder for reconstruction by masking out random areas on the input point set. 

\subsection{Conventional Knowledge Distillation. }
Aiming at improving the performance of deep learning models, a common way is to average all predictions on the same data from multiple models to reduce model variance and generalization error. This cumbersome scheme expedites the invention of conventional knowledge distillation that transfers the knowledge among different models~\cite{hinton2015distilling}. Generally, teacher networks with greater model complexity learn knowledge and guide student networks with more compactness to learn from better representations for enhanced generalization. As a result, student networks are more suitable for deployment on systems with less computational resource~\cite{blakeney2020parallel, li2021online}, e.g., mobile phone and circuit board. Other investigations into knowledge distillation, with an emphasis on early stopping of teacher's training, have asserted that fully converged teachers do not typically promote student learning~\cite{wang2022efficient, cho2019efficacy}. In this study, we trivially bypass the need for optimal teacher selection through end-to-end self-distillation.
\begin{figure*}[t]
\begin{center}
    \includegraphics[width=1.0\linewidth]{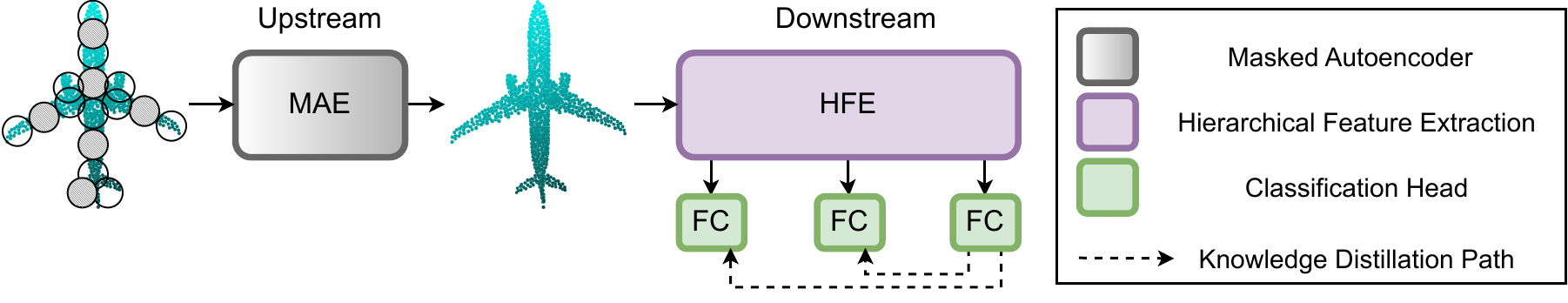}
\end{center}
   \caption{The architecture of the proposed cascaded network. The upstream network functions as a masked autoencoder, which reconstructs the incomplete input point cloud into a complete shape. The downstream network is comprised of a hierarchical feature extraction (HFE) module and fully connected (FC) classification heads. 
   The flow of knowledge is represented by the dash lines  $\dashrightarrow$, with the last level serving as the teacher to guide students in the former levels.}
\label{fig:arch}
\end{figure*}

\subsection{Self-Distillation. }
Born-again networks~\cite{furlanello2018born} revisited knowledge distillation with a different objective by disentangling the training procedure from model compression. In a progressive fashion, the student models that are identical to the teacher model are trained consecutively from the supervisions of previous generations, forming an ensemble at the end that even outperforms the teacher model at the first generation. 

Recently, a new concept called self-distillation has emerged following the development of born-again networks. This can be divided into two categories based on whether knowledge is distilled within~\cite{chen2022sdae, zhang2021self, zhang2021perturbed} or out of~\cite{kim2021self, zhang2020self} the same iteration in the training stage, for tasks such as label smoothing or weakly supervised learning. Our proposed method belongs to the former one, in which we train each classification model simultaneously regardless of the roles. Inspired by those born-again networks,
we tend to transfer the knowledge from the deepest level of the hierarchy to shallower models through self-distillation.

\subsection{Point Cloud Understanding. }
Studies on point cloud have demonstrated impressive performance across widely recognized benchmarks. Pioneered by PointNet~\cite{qi2017pointnet}, various hierarchical scaling networks have been proposed to enhance both global and local feature awareness, including PointNet++~\cite{qi2017pointnet2}, DGCNN~\cite{wang2019dynamic}, CurveNet~\cite{xiang2021walk}, PRA-Net~\cite{cheng2021net}, MKANet~\cite{xiao2024mkanet}, PointMLP~\cite{ma2022rethinking}, PointRas~\cite{zheng2022pointras}, Point-PN~\cite{zhang2023starting}, APP-Net~\cite{lu2023app}, SPoTr~\cite{park2023self}, PointConT~\cite{liu2023point}, GLSNet++~\cite{bao2024glsnet++} and 	
PointGST~\cite{zhou2024dynamic}. Networks such as PointNet++ and DGCNN establish hierarchical structures by capturing point relationships through multi-scale grouping (MSG) or k-nearest neighbors (kNN) within local regions. 
CurveNet traverses sequences of connected points, creating hypothetical curves that enrich pointwise features, while PointMLP reinterprets PointNet++ with simple yet effective residual connections, achieving state-of-the-art results.
In this study, we incorporate several well-established hierarchy-based networks (i.e., PointNet++, CurveNet, PointMLP) in our joint learning framework to support downstream tasks.

\section{Method}
\subsection{Preliminary}
\noindent\textbf{Problem Formulation. }
Given the complete point set $Q = \{ q_i | i=1,...,N\} \in \mathbb{R}^{N \times 3}$ and its incomplete counterpart $X' = \{ x_i' | i=1,...,N'\} \in \mathbb{R}^{N' \times 3}$, we seek for a solution to reconstruct $Q$ from $X'$ and simultaneously classify or segment the reconstructed $X = \{ x_i | i=1,...,N\} \in \mathbb{R}^{N \times 3}$, where $N'$ and $N$ denote the number of each point cloud within  Euclidean space. The architecture of our method is illustrated in Fig.~\ref{fig:arch}.
Our end-to-end architecture is composed of upstream and downstream sub-networks, denoted as $\mathcal{F}_u$ and $\mathcal{F}_d$, respectively.

\noindent\textbf{Information Bottleneck. }
Let the random variable $X$ represent the input data, $Y$ represent the output/target labels, the information bottleneck (IB) principle~\cite{tishby2000information} aims to find a compressed representation of $X$, denoted by $Z$, that minimizes the mutual information $I(X;Z)$ and maximizes the relevant information $I(Y;Z)$. This objective can be formulated as:
\begin{equation}\label{eq:ib}
\scalemath{0.86}{
    \underset{Z}{\min} \{ I(X;Z) - \beta I (Y;Z) \},
}
\end{equation}
where $\beta$ is a Lagrange multiplier that constrains the expected distortion.

\noindent\textbf{Strong Regularization. }
To solve the gradient vanishing problem, Deeply-Supervised Nets (DSN)~\cite{lee2015deeply} simultaneously minimizes loss functions across $L$ recognition layers~\cite{li2022comprehensive}. DSN acts as a strong regularization, smoothing gradients during backpropagation. From the perspective of IB principle, DSN optimization can be formulated as:
\begin{equation}\label{eq:dsn} 
\scalemath{0.86}{
    \underset{Z}{\min} \{I(X;Z_L) - \beta I (Y;Z_L) + \underbrace{\sum_{l=1}^{L-1} [I(X;Z_l) - \beta I (Y;Z_l)]}_{\text{reg: }\mathcal{R}_\mathit{DSN}}\},
}
\end{equation}
where $\beta$ controls the emphasis on preserving relevant information.
In a cascaded manner, each $l \in [1,2,...,L]$ layer functions analogously to an ensemble of previous layers, referred to as companions.
Notably, the second term serves as a regularization over a conventional single objective function. Intuitively, DSN treats both the output layer and intermediate companion layers equally. 

\begin{figure*}[h!]
\begin{center}
    \includegraphics[width=1.0\linewidth]{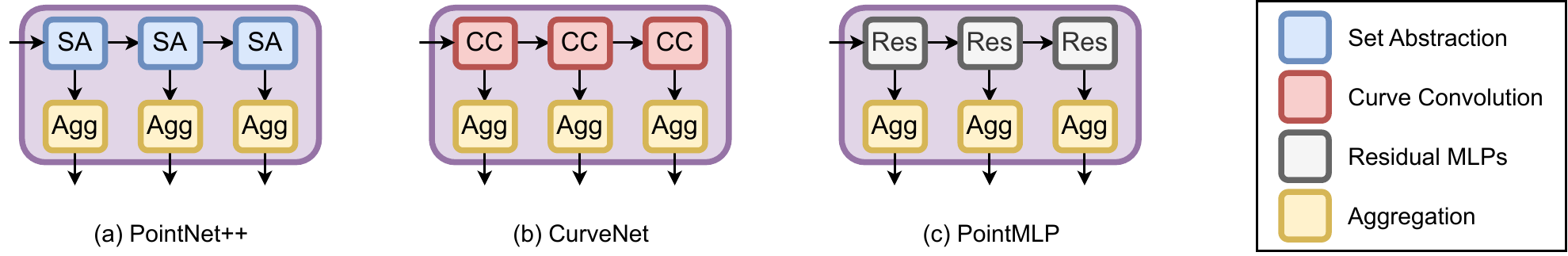}
\end{center}
   \caption{Internal structures of hierarchical feature extraction modules.
   }
\label{fig:hfe}
\end{figure*}

\subsection{Upstream Network}\label{sec:rec}
In real scenes, scanned point clouds are often incomplete or imperfect due to factors such as sensor resolution limitations, obstacles, or vibrations. To alleviate this phenomenon, we adopt a simplified version of SnowflakeNet as our upstream network $\mathcal{F}_u$, which learns the mapping between $Q$ and $X$ and is optimized using a reconstruction loss: 
\begin{equation}\label{eq:rec}
\scalemath{0.86}{
\mathcal{L}_\mathit{rec} = \frac{1}{|X|} \sum_{x \in X} \underset{q \in  Q}{\min} {||x-q||}_2^2 + \frac{1}{|Q|} \sum_{q \in Q} \underset{x \in  X}{\min} {||q-x||}_2^2,
}
\end{equation}
where ${||\cdot||}_2^2$ denotes the squared Euclidean 2-norm of Chamfer distance (CD). This objective in Eq.~\ref{eq:rec} quantifies the bidirectional similarities between the two point sets, promoting the reconstructed point cloud's proximity to the target point cloud.

\subsection{Downstream Network}\label{sec:downnet}
Generally, the downstream network is dependent upon varying tasks. Targeting on point cloud classification problem, we employ three elite versions of canonical hierarchy-based methods, i.e., PointNet++, CurveNet, and PointMLP, as our downstream networks, each of which has demonstrated satisfactory local awareness. The objective for this task obeys Eq.~\ref{eq:dsn}, defined by the cross-entropy loss, denoted as $\mathcal{L}_\mathit{ce}$. Without losing generality, most point cloud processing methods incorporate aggregation operations (e.g., max pooling, summation, and global set abstraction) at the end of feature extraction to accommodate the orderless nature of point clouds. However, these operations may unintentionally eliminate distinguishable features, leading to degraded performances. To address this, we propose a hierarchical self-distillation framework for feature fusion across different scales, as described in the following section.

\subsection{Stronger Regularization}\label{sec:hsd}

Hierarchy-based networks generate multi-level features derived from various scales or levels, e.g., the integrated features from MSG of PointNet++~\cite{qi2017pointnet2} and the intermediate features from multihead attentions of GSRFormer~\cite{cheng2022gsrformer} or from geometry-based multiprojection fusion modules of \cite{xu2023multi}. However, MSG-like schemes have limitations, as they do not capture the dissociated information among features. Intuitively, an incomplete object might be encoded incorrectly if it lacks a critical part essential for its representation. Hence, our approach involves constructing multiple classification heads for hierarchical feature extraction, coupled with self-distillation to capture comprehensive knowledge across all levels. We term this approach \textit{PointHSD}, a general framework applicable to any hierarchy-based point cloud method. A universal structure for hierarchical feature extraction (HFE) that includes the elite versions of classical point cloud classification networks (i.e., PointNet++, CurveNet, and PointMLP) is depicted in Fig.~\ref{fig:hfe}. Here, set abstraction, curve convolution, and residual MLPs act as local feature extractors, incorporating with local aggregation layers in each module. 
 
The Hierarchical Feature Extraction (HFE) module, as shown in Fig.~\ref{fig:arch}, is responsible for learning high-level local details from the low-level reconstructed point set $X$. This is achieved by using different branches, each representing a distinct range of surroundings, i.e., by using $k_1$, $k_2$, and $k_L$ to extract kNN features in the branches. The outputs of these branches are aligned to identical dimensions through 3 independent fully-connected layers, facilitating subsequent distillation processes.
Let $\mathcal{F}_d = \coprod_l^L \mathcal{F}_l$ denote this procedure, which generates intermediate and deepest latent representations $Z_l$ and $Z_L$, respectively. We expect $Z_L$ to have a larger perceptual field, such that the difference between $Z_L$ and the intermediate representations $Z_l$ can be effectively captured through self-distillation. The regularization term $\mathcal{R}_\mathit{DSN}$ in Eq.~\ref{eq:dsn} is thus reformulated as:
\begin{equation}\label{eq:hsd}
\scalemath{0.86}{
    \sum_{l=1}^{L-1} [I(X;Z_l) - \beta \gamma I (Y;Z_l) - \underbrace{ \beta (1-\gamma) I (Z_L;Z_l)]}_{\text{more reg: }\mathcal{R}_\mathit{HSD}},
}
\end{equation}
where by maximizing the mutual information $I(Z_L;Z_l)$, we can equivalently quantify the difference (denoted as $\mathcal{L}_\mathit{kl}$) by minimizing the Kullback-Leibler (KL) divergence. Accordingly, the third term $\mathcal{R}_\mathit{HSD}$ in Eq.~\ref{eq:hsd} can be implemented as $\mathcal{L}_\mathit{kl}$ to further regularize the optimization by smoothing the hard predicted labels of students.
DSN in Eq.~\ref{eq:dsn} aims to enhance feature quality across all hidden layers, functioning as a type of feature regularization to alleviate overfitting~\cite{lee2015deeply}. During backpropagation, all companion outputs are treated equivalently, ensuring that features from different hidden layers are associated to the common proxy, rather than solely with the final layer. In contrast, the proposed HSD approach not only emphasizes the collective contributions of all companions to enhance feature quality but also assigns a supervisory role to the last layer, acting as a ``teacher" to guide preceding, potentially uncertain, companions toward a more refined representation by maximizing the mutual information. This mechanism allows the last layer to provide more informative guidance to earlier layers, alleviating overfitting issues in companion layers. More intuitively, this strategic supervision aims to support the iterative refinement of the classification process, especially in scenarios where multiple attempts at classification are needed to achieve an accurate representation of a roughly reconstructed shape.
Therefore, our HSD framework distinguishes between the output layer and its companions. Notably, deep supervision can be seen as a special case of Eq.~\ref{eq:hsd} if $\gamma=1$. 

Mathematically, as $X$  may still lack certain regions, the reconstructed shape could contain unrelated or less relevant features, such that $I(Y;Z_l) \leq I(Y;Z_L)$. For any $l < L$, let $k_l < k_L$ be the $k$ nearest neighbors for feature extraction in each layer. As $k$ increases, the mutual information between the class label and the latent feature also increases, reflecting a more accurate representation of the 3D point cloud due to the integration of broader neighborhood information.
Hence, $\mathcal{R}_\mathit{HSD}$ in Eq~\ref{eq:hsd} facilitates the transfer of broader contextual features from layer $L$ to flow back to previous ones. 
Alongside the cross-entropy term associated with student layers, the term $\mathcal{R}_\mathit{HSD}$ imparts smoothness to the probabilistic distribution through the convolution of information derived from the teacher. This mechanism contributes to \textbf{stronger regularization} of the students, thereby ultimately mitigating the overconfidence of the ensemble model.

\subsection{Joint Learning}\label{sec:jl}

Recall that as the resolution or completeness of input point clouds $X'$ declines, classification accuracy tends to decline significantly, primarily because incomplete shapes can yield misleading local features.

In order to overcome this issue, we introduce a straightforward yet effective function to jointly learn both object completeness and category, defined as:
\begin{equation}
\scalemath{0.86}{
    \mathcal{J} = \mathcal{F}_u \circ \mathcal{F}_d = \mathcal{F}_u \circ \coprod_l^L \mathcal{F}_l,
}
\end{equation}
where $\circ$ denotes the cascade of two streams, and $\mathcal{F}_d$ follows the Markov chain with the assumption that $I (Y, Z_1) \leq I (Y, Z_2) \leq ... \leq I (Y, Z_L)$. Therefore, we designate the deepest layer as the teacher, as it encapsulates  most comprehensive knowledge to soften hard labels in the other layers. The mutual information $I (X, Z_l)$ is implicitly compressed by the network. Finally, combining Eqs.~\ref{eq:dsn}, ~\ref{eq:rec}, and ~\ref{eq:hsd}, we optimize our joint learning objective as follows: 
\begin{equation}\label{eq:joint}
\scalemath{0.86}{
\mathcal{L}_\mathit{joint} = \alpha \mathcal{L}_\mathit{rec} + \beta \gamma \mathcal{L}_\mathit{ce} + \beta (1-\gamma) \mathcal{L}_\mathit{kl},
}
\end{equation}
where $\alpha$, $\beta$, $\gamma$ are the weights, which are empirically set to 1, 0.001, and 0.8, respectively. $\alpha$ and $\beta$ control the balance between upstream and downstream processes, which are guided by the sparse data-focused paradigm design that requires reconstruction as a prerequisite. Additionally, $\beta$ distorts the amount of information used for direct optimization in classification or for flowing back to regularize the network. In this way, our network learns to sequentially and simultaneously reconstruct and recognize scattered point clouds, merging both geometric and semantic priors into the model.

It is important to note that our joint learning framework shares similarities with the cascaded fusion network, CasFusionNet~\cite{xu2023casfusionnet}, which is designed for scene completion and semantic segmentation. However, a fundamental distinction lies in that CasFusionNet focuses on dense feature fusion within the hyperspace, amalgamating both geometric and semantic features, while our method extends its influence by integrating both low-level and high-level features to achieve a synergistic enhancement across tasks. Moreover, unlike CasFusionNet, the plug-and-play property of our framework allows either the upstream or downstream network to be replaced to better suit specific application needs.

\section{Experiment}\label{sec:exp}

\noindent\textbf{Dataset.}
Our data are sourced from three widely recognized benchmarks: ModelNet40~\cite{wu20153d}, ScanObjectNN~\cite{uy2019revisiting}, and ShapeNetPart~\cite{yi2016scalable}. ModelNet40, a synthetic mesh dataset, includes 9,843 training and 2,468 testing samples, classified into 40 categories. In contrast, ScanObjectNN presents a challenging real-world dataset comprising 15,000 objects across 15 categories, including 2,902 unique instances. To generate scattered point clouds, we employ a sampling strategy that captures representative patches: instead of randomly masking out points, we use farthest point sampling (FPS) to select patch centroids, then incorporate surrounding points via the kNN algorithm. For ModelNet40, we prepare three input types with the following sizes: $32 \times 16=512$, $32 \times 8=256$, and $64 \times 8=512$ , where the latter configuration uses eight neighbors sampled at a denser resolution from the ground truth (GT), making it much sparser than the other configurations. For ScanObjectNN and ShapeNetPart, we randomly sample $64 \times 16$ points as input.

\begin{table}[t]
\begin{center}
\caption{Details of the inputs with various sparsities.}\label{tab:data}
\begin{tabular}{l|cc}
\hline
\rowcolor{gray!20}
Dataset & Resolution & Sparsity \\
\hline
\multirow{3}{*}{ModelNet40} & $32 \times 16$ & Low \\
 & $32 \times 8$ & Medium \\
 & $64 \times 8$ & High \\
\hline
\multirow{2}{*}{ScanObjectNN} & $64 \times 16$ & Medium \\
& $128 \times 8$ & Medium \\
\hline
ShapeNetPart & $64 \times 16$ & Medium \\
\hline
\end{tabular}
\end{center}
\end{table}
\begin{figure}[t]
\begin{center}
    \includegraphics[width=.9\linewidth]{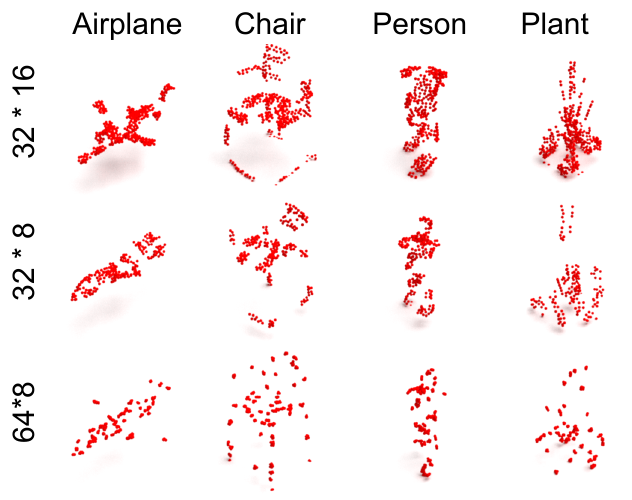}
\end{center}
   \caption{Different types of input data on ModelNet40. From top to bottom, each local area has less local range coverage, resulting in higher sparsity.}
\label{fig:input}
\end{figure}

\begin{table}[t]
\caption{Classification results of downstream-only networks on ModelNet40 with full inputs. The proposed HSD on all three baseline models have improved both OA and mAcc. These results are obtained on a single Nvidia 3090 GPU.}\label{tab:base}
\begin{center}
\begin{tabular}{lcc}
\hline
\rowcolor{gray!20}
Model & OA(\%) & mAcc(\%) \\
\hline
PointNet++~\cite{qi2017pointnet2} & 91.8 & 89.2 \\
CurveNet~\cite{xiang2021walk} & 92.5 & 89.4 \\
PointMLP~\cite{ma2022rethinking} & 92.7 & 89.7 \\
\hline
\rowcolor{LightCyan}
PointNet++-HSD (Ours) & \textbf{92.6} & \textbf{89.5} \\
\rowcolor{LightCyan}
CurveNet-HSD (Ours) & \textbf{92.6} & \textbf{90.0} \\
\rowcolor{LightCyan}
PointMLP-HSD (Ours) & \textbf{93.2} & \textbf{90.3} \\
\hline
\end{tabular}
\end{center}
\end{table}

\begin{table*}[t]
\caption{Performance comparison between SOTA two-stage methods (the name of the upstream model, SnowflakeNet, is omitted) and our joint learning ($\mathcal{J}$) methods on ModelNet40 with respect to three types of sparsity. The self-distillation to all three baseline models has led to improved overall classification accuracy while preserving competitive reconstruction errors.}\label{tab:incompcls}
\begin{center}
\begin{tabular}{l|ccc|ccc|ccc}
\hline
\rowcolor{gray!20} & \multicolumn{3}{c}{$32 \times 16$} & \multicolumn{3}{c}{$32 \times 8$} & \multicolumn{3}{c}{$64 \times 8$ (Sparser)}  \\
\cline{2-10}
\rowcolor{gray!20}
\multirow{-2}{*}{Model} & OA(\%) & mAcc(\%) & CD & OA(\%) & mAcc(\%) & CD & OA(\%) & mAcc(\%) & CD \\
\hline
PointNet++~\cite{qi2017pointnet2} & 89.0 & \textbf{88.7} & \multirow{5}{*}{23.04} & 87.0 & \textbf{86.6} & \multirow{5}{*}{\textbf{27.33}} & 70.3 & 70.1 & \multirow{5}{*}{40.78} \\
DGCNN~\cite{wang2019dynamic} & 79.9 & 72.4 & & 77.6 & 71.8 & & 65.3 & 57.6 & \\
PCT~\cite{guo2021pct} & 88.9 & 85.0 & & 86.6 & 81.6 & & 71.7 & 64.8 & \\
CurveNet~\cite{xiang2021walk} & 84.6 & 79.1 & & 82.1 & 76.0 & & 69.0 & 60.0 &  \\
PointMLP~\cite{ma2022rethinking} & 85.7 & 78.5 & & 81.7 & 72.9 & & 66.9 & 59.6 & \\
\hline
Our $\mathcal{J}$ (PointNet++) & 89.9 & 86.2 & 23.22 & 89.1 & 85.0 & 27.41 & 82.7 & \textbf{77.4} & \textbf{38.10} \\
Our $\mathcal{J}$ (CurveNet) & 89.0 & 86.0 & \textbf{23.03} & 88.8 & 85.8 & 27.79 & 80.4 & \textbf{76.1} & \textbf{38.71} \\
Our $\mathcal{J}$ (PointMLP) & 89.9 & 87.1 & 23.15 & 89.4 & 85.6 & 27.49 & 79.7 & 75.7 & 39.67 \\
\rowcolor{LightCyan}
Our $\mathcal{J}$ (PointNet++-HSD) & \textbf{90.8} & 87.7 & 23.10 & \textbf{89.9} & \textbf{86.6} & 27.79 & \textbf{82.9} & 76.9 & 39.10 \\
\rowcolor{LightCyan}
Our $\mathcal{J}$ (CurveNet-HSD) & \textbf{89.1} & 85.5 & 23.16 & \textbf{88.8} & 85.3 & 28.02 & \textbf{81.3} & 75.5 & 39.17 \\
\rowcolor{LightCyan}
Our $\mathcal{J}$ (PointMLP-HSD) & \textbf{90.3} & 87.0 & 23.27 & \textbf{88.9} & 85.0 & 28.36 & \textbf{81.8} & \textbf{75.8} & \textbf{39.42} \\
\hline
\end{tabular}
\end{center}
\end{table*}

\noindent\textbf{Data Preparation. }
We sample scattered point clouds based on a specific number of seeds and neighbors, leveraging FPS and kNN to generate varied sparsity levels. Table~\ref{tab:data} details the sparsity configurations used for each dataset, and Fig.~\ref{fig:input} illustrates the sparsity variations in inputs for ModelNet40. Specifically, the configuration  with $64 \times 8 = 512$ points has the most scattered distribution, as neighbors are selected from a dense point set of 10,000 points. This creates a higher sparsity type, where each cluster has limited range coverage yet high local density. 
Our sparse sampling methodology inherently mirrors random masking but operates additively rather than by reduction. This approach emulates a continuous scanning scenario, simulating an object being progressively scanned from various perspectives with potential occlusions. Unlike random masking, which simply omits regions, this method emphasizes a more focused understanding of single objects, making it well-suited for understanding tasks.
The data pre-processing code for all datasets are provided within our source code.

\noindent\textbf{Implementation Detail.}
Our networks are trained on a server deployed with 10 Nvidia RTX 2080Ti GPUs, using a batch size of 260 for classification tasks and 160 for segmentation tasks across all models, both during training and testing phases. In the downstream, an upsampling ratio of (1, 1, 2) is applied by default to point-splitting modules. 
For consistency, we omit the upstream network name (i.e., SnowflakeNet) and directly use the downstream model names to denote two-stage methods; we refer to our joint learning framework as \textbf{$\mathcal{J}$}. Take our PointNet++-HSD for instance, we configure $k_1=8, k_2=16, k_L=24$ to gradually increase coverage ranges. To maintain experimental simplicity, we do not apply a voting strategy in any experiments. Our source code is available at: \url{https://github.com/ky-zhou/PointHSD}.

\noindent\textbf{Evaluation Metrics.} In line with standard practices in the literature, we report the following metrics:
Overall accuracy (OA), mean-class accuracy (mAcc), and Chamfer distance (CD) ($\times 1000$) for classification, and mean-instance and mean-class intersection over union (mIOU/cIOU) for segmentation.

\subsection{Classification with Self-Distillation}\label{sec:exp-sd}

In this study, we evaluate the classification performance of PointNet++, CurveNet, and PointMLP as baseline models, comparing their accuracy against our proposed Hierarchical Self-Distillation (HSD) framework. To underscore the impact of HSD on classification, we test each downstream network with full point set inputs, as shown in Table~\ref{tab:base}. The results indicate that HSD enhances classification outcomes by refining feature quality across layers. Since the input point cloud is complete, only a single viewpoint is used for this task. For fair comparison, we re-train all baseline models on the same machine and with identical data splits as our HSD-based models. Following standard community practices, we use 1,024 input points. The highest testing results for OA and mAcc are reported. To regulate the learning rate, we employ a cosine annealing scheduler with a minimum rate of 0.005 as same as PointMLP. It is noteworthy that HSD for complete shapes operates similarly to DSN as the feature quality is typically sufficient after the initial extraction layer. Optimal performance in these cases may arise from intermediate layers, which aligns with the understanding that, for complete objects, accuracy hinges on the selection of local kernel coverage at each layer. This suggests that HSD may be particularly adept at handling scattered point clouds. In the following section, we present quantitative results that demonstrate the effectiveness of HSD applied to sparse data.

\subsection{Incomplete Point Cloud Classification}\label{sec:exp-ipc}

\begin{table}
\caption{Performance comparison between the SOTAs (upstream names omitted) and our joint learning ($\mathcal{J}$) methods on ScanObjectNN's variant (PB\_T50\_RS: object w. BG) with $64 \times 16$ scattered points as input. Our proposed model concretely outperforms existing methods.}\label{tab:scannn2}
\begin{center}
\begin{tabular}{l|ccc}
\hline
\rowcolor{gray!20}
Model & OA(\%) & mAcc(\%) & CD \\
\hline
PointMAE~\cite{pang2022masked} & 77.6 & 74.7 & \multirow{4}{*}{17.79} \\
PointMLP~\cite{ma2022rethinking} & 77.7 & 75.1 &  \\
Point-PN~\cite{zhang2023starting} & 79.0 & 76.5 &  \\
PointConT~\cite{liu2023point} & 81.2 & 78.1 &  \\
\hline
Our $\mathcal{J}$ (PointNet++) & 79.1 & 76.1 & 17.95 \\
Our $\mathcal{J}$ (CurveNet) & 80.1 & 77.0 & 17.76 \\
Our $\mathcal{J}$ (PointMLP) & 81.6 & 78.9 & 17.82 \\
\rowcolor{LightCyan}
Our $\mathcal{J}$ (PointNet++-HSD) & \textbf{81.3} & \textbf{79.0} & \textbf{16.79} \\
\rowcolor{LightCyan}
Our $\mathcal{J}$ (CurveNet-HSD) & \textbf{81.6} & \textbf{78.8}  & \textbf{16.74} \\
\rowcolor{LightCyan}
Our $\mathcal{J}$ (PointMLP-HSD) & \textbf{82.7} & \textbf{80.8} & \textbf{16.35} \\
\hline
\end{tabular}
\end{center}
\end{table}

\begin{figure*}[t]
\centering
    \begin{subfigure}[b]{0.32\linewidth}
        \includegraphics[width=\textwidth]{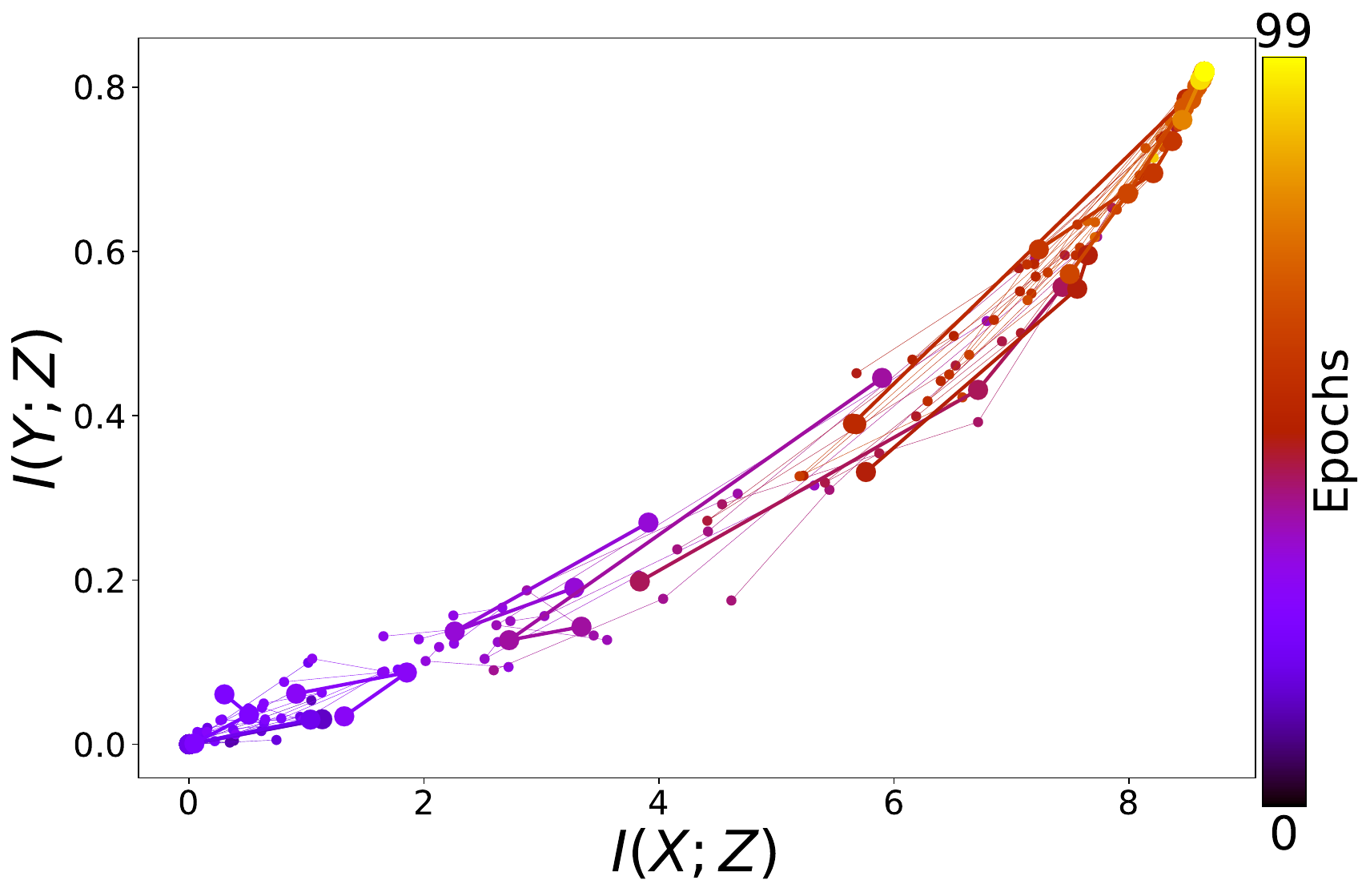}
        \caption{Information flow}\label{fig:infop}
    \end{subfigure}
\hfill
    \begin{subfigure}[b]{0.32\linewidth}
        \includegraphics[width=\textwidth]{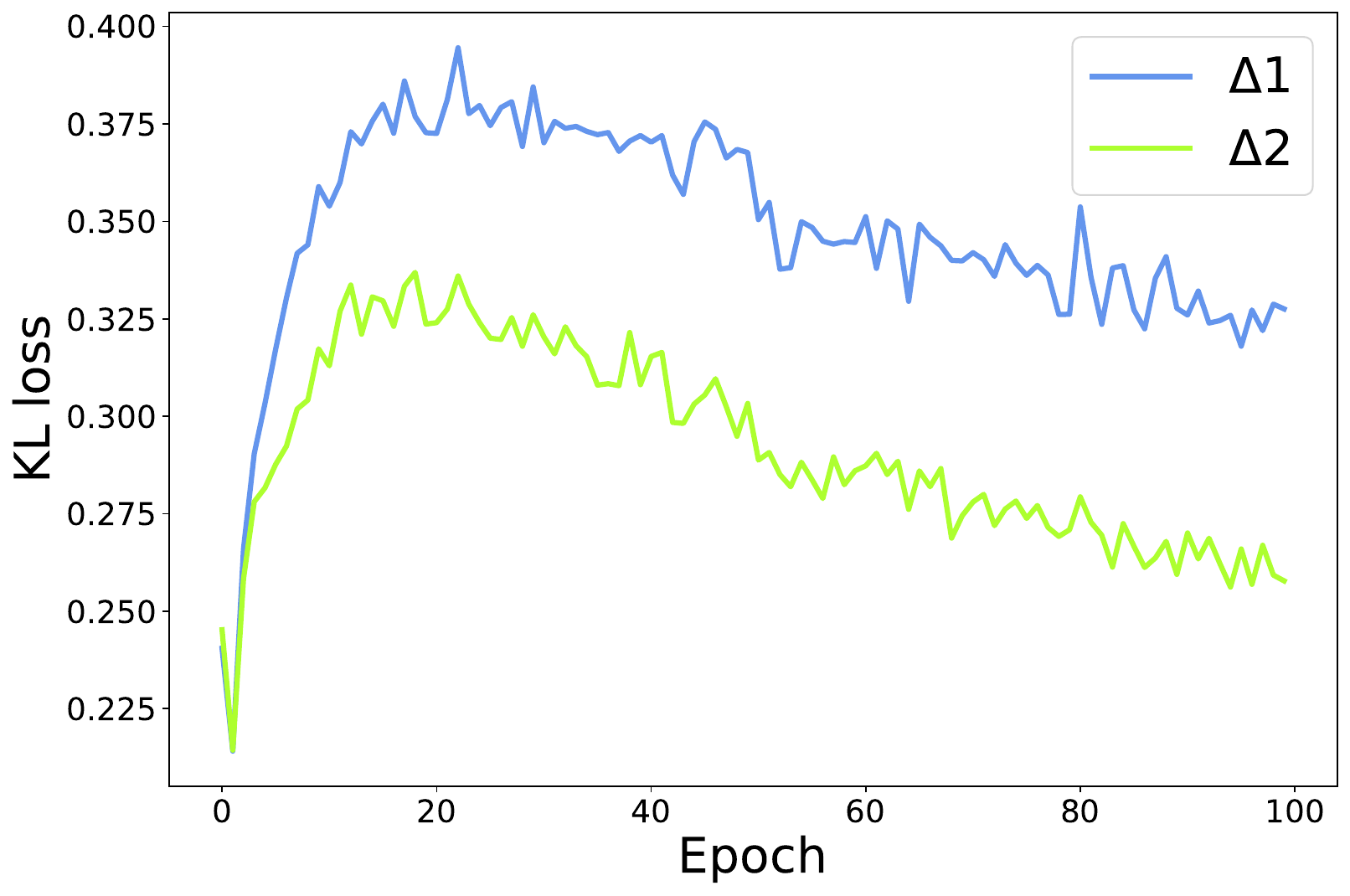}
        \caption{KL}\label{fig:kl}
    \end{subfigure}
\hfill
    \begin{subfigure}[b]{0.32\linewidth}
        \includegraphics[width=\textwidth]{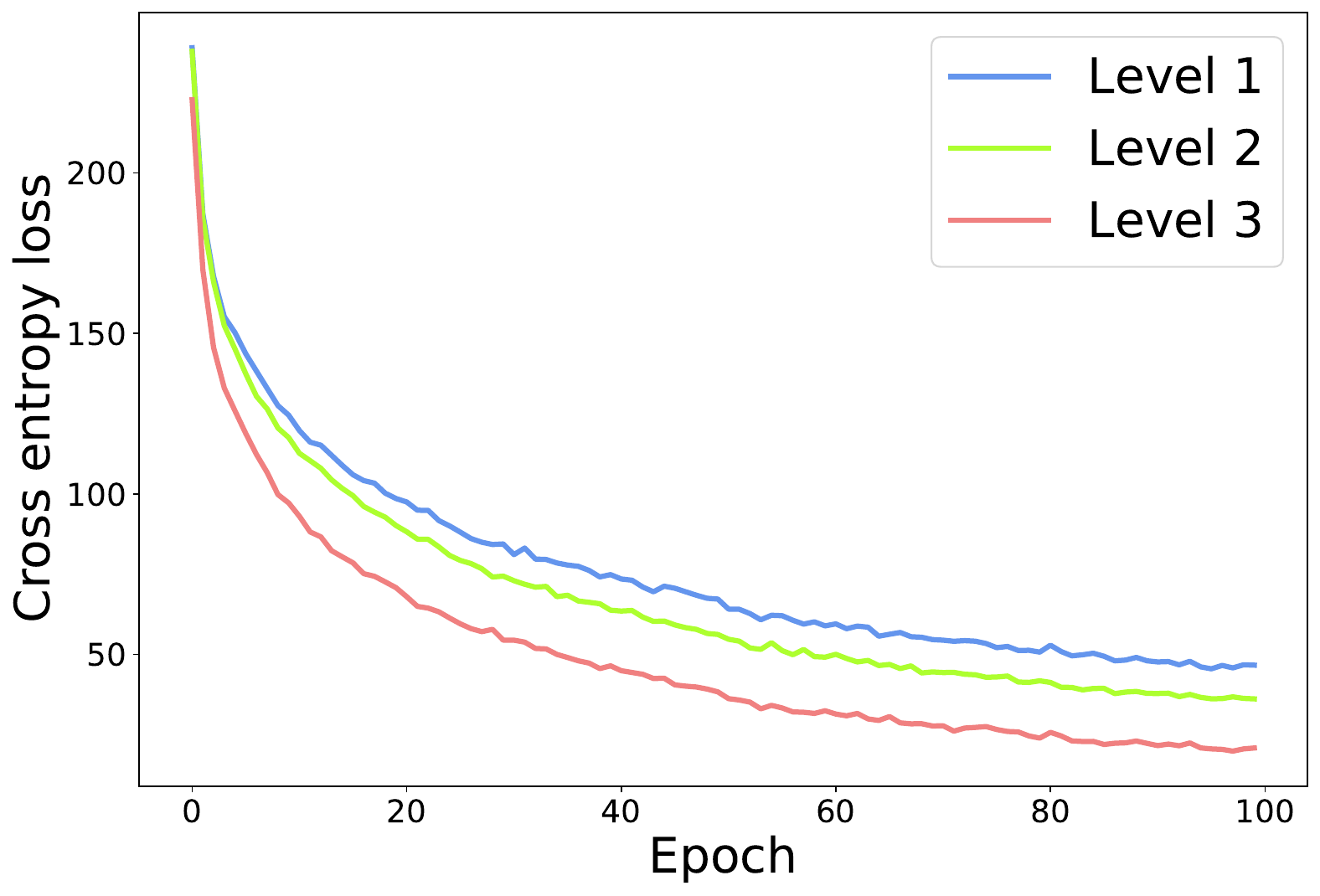}
        \caption{Cross entropy}\label{fig:ce}
    \end{subfigure}
\caption{Quantitative visualizations. (a) While difference among 3 levels is visually subtle, the information across all levels progressively increases, ultimately reaching a maximum. Larger circles are emphases every specific intervals. (b) The differences are quantified by KL loss, comparing level 1 and level 3 ($\Delta 1$) and level 2 and level 3 ($\Delta 2$). (c) Mutual information is represented by cross-entropy loss, where level 3 is the teacher.}\label{fig:quan_ana}
\end{figure*}

\begin{table*}[h]
\caption{Evaluation results of our joint learning ($\mathcal{J}$) methods on ScanObjectNN's variant (PB\_T50\_RS: object w/o. BG).}\label{tab:scannn1}
\begin{center}
\begin{tabular}{l|ccc|ccc}
\hline
\rowcolor{gray!20} & \multicolumn{3}{c}{$64 \times 16$} & \multicolumn{3}{c}{$128 \times 8$} \\
\cline{2-7}
\rowcolor{gray!20}
\multirow{-2}{*}{Model} & OA(\%) & mAcc(\%) & CD & OA(\%) & mAcc(\%) & CD \\
\hline
SnowflakeNet~\cite{xiang2021snowflakenet} & - & - & 12.00 & - & - & 11.05 \\
\hline
Our $\mathcal{J}$ (PointNet++) & 82.2 & 80.1 & 12.11 & 82.2 & 79.8 & 11.12 \\
Our $\mathcal{J}$ (CurveNet) & 82.8 & 80.5 & 12.03 & 83.1 & 80.9 & 11.10 \\
Our $\mathcal{J}$ (PointMLP) & 84.5 & 82.8 & 12.11 & 84.2 & 82.5 & 11.16 \\
\rowcolor{LightCyan}
Our $\mathcal{J}$ (PointNet++-HSD) & \textbf{84.5} & \textbf{82.7} & \textbf{11.16} & \textbf{84.9} & \textbf{83.5} & \textbf{9.98} \\
\rowcolor{LightCyan}
Our $\mathcal{J}$ (CurveNet-HSD) & \textbf{83.7} & \textbf{81.3} & \textbf{11.97} & \textbf{83.6} & \textbf{80.9} & \textbf{10.04} \\
\rowcolor{LightCyan}
Our $\mathcal{J}$ (PointMLP-HSD) & \textbf{85.2} & \textbf{83.3} & \textbf{11.49} & \textbf{85.1} & \textbf{83.3} & \textbf{10.17} \\
\hline
\end{tabular}
\end{center}
\end{table*}

Point sets captured by real sensors, such as LiDAR, are typically distributed only on an object's surface, predominantly on the side facing the sensor, which often leads to classification misalignment and challenges in boundary detection.  The upstream network $\mathcal{F}_\mathit{u}$ tackles this challenge by reconstructing the incomplete point cloud into a complete shape. In this task, all of our models process incomplete point clouds as input. All two-stage models are built with \textbf{full} architectures, while our downstreams are \textbf{elite} versions.
The target point cloud size is set to 1,024 for fair comparisons. During training, each incomplete input is augmented with 8 different views. Additionally, we apply a gradual learning rate decay by a factor of 0.8 every 50 epochs, starting with an initial learning rate of 0.001. In following experiments, we denote $\mathcal{J}$ as our joint learning methods.

\noindent\textbf{Results on ModelNet40. }
The results in Table~\ref{tab:incompcls} reflect that the proposed HSD models outperform the baselines in terms of OA for incomplete object classification. We also conduct completion and classification tasks separately rather than in an end-to-end scheme. In particular, the results generated by SnowflakeNet are directly utilized as inputs to fully trained downstream classification networks. For PointNet++, given the variability in predictions, we perform the evaluation four times and report the average results for both OA and mAcc. Our HSD models achieve the highest OA with competitive CD compared to two-stage methods. Notably, our proposed methods significantly outperform two-stage approaches on sparsely masked data.
We infer that our joint learning models enhance the information gain especially on sparsely distributed data, mimicking the inference procedure in the real-world scenarios. It is also observed that HSD is less effective when applied with CurveNet. 

\noindent\textbf{Results on ScanObjectNN. }
In this task, we follow the standard in the community and use the hardest variant of ScanObjectNN: PB\_T50\_RS with background. In Table~\ref{tab:scannn2},
although our baseline models (i.e., without HSD) have already achieved competitive performances, our HSD models demonstrate superior results in terms of OA, mAcc, and CD, outperforming SOTA methods that are specifically trained on complete shapes from ScanObjectNN. Notably, all of our jointly learned HSD models surpass the most-recent hierarchy-based methods Point-PN~\cite{zhang2023starting} and PointConT~\cite{liu2023point}, underscoring the effectiveness and relevance of our approach in real-world scenarios. More results on ScanObjectNN without background are given in Table.~\ref{tab:scannn1}. Here, we primarily  compare our methods to our baselines, except for the upstream's reconstruction error without joint learning. Once again, our proposed methods achieve lower CD compared to using a standalone completion network like SnowflakeNet.

\noindent\textbf{Why HSD works less effective on CurveNet? }
Empirically, despite CurveNet has a hierarchical nature, it walks among the points, forming the curves to represent local features. Unlike query ball and kNN commonly used in traditional hierarchical methods, curves share more discriminative features in the same region, resulting in less mutual information shared among different levels.

\subsection{Quantitative Analysis}

\begin{figure*}[ht]
\begin{center}
    \includegraphics[width=1.0\linewidth]{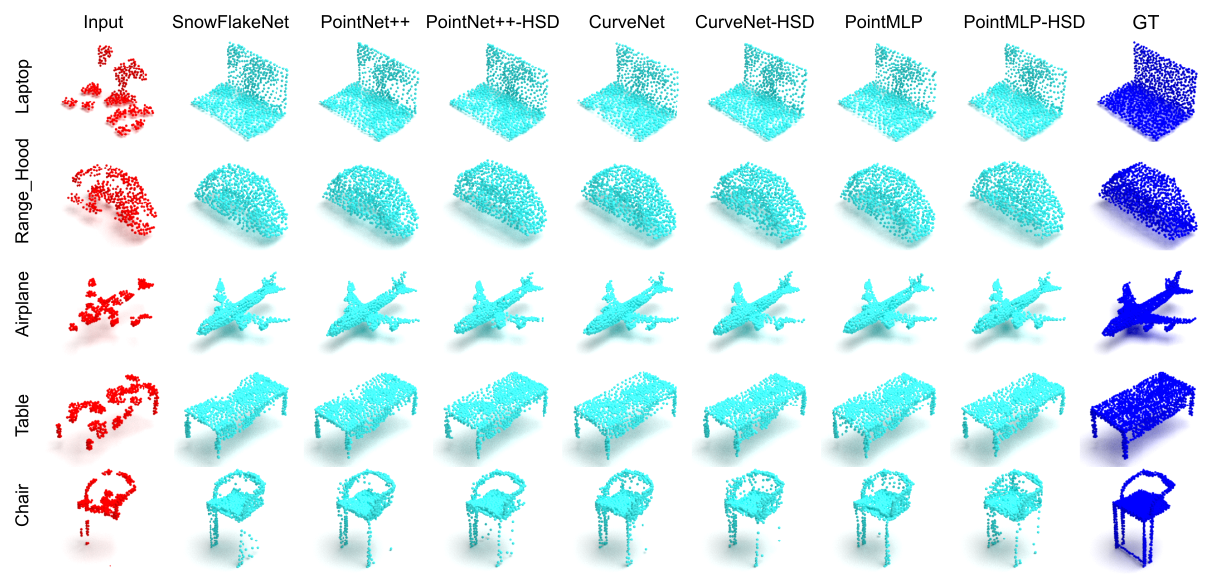}
\end{center}
   \caption{Completed objects of upstream on ModelNet40.}
\label{fig:recon1}
\end{figure*}
\begin{figure}[h]
\begin{center}
   \includegraphics[width=1.0\linewidth]{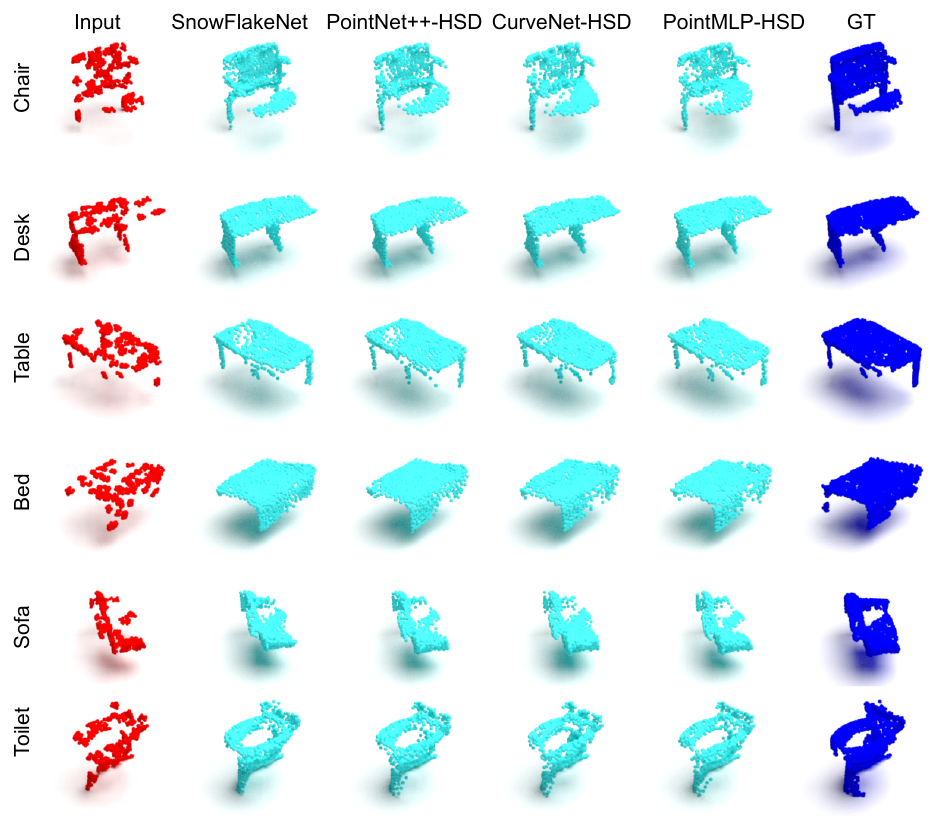}
\end{center}
   \caption{Completed objects of upstream on ScanObjectNN.}
\label{fig:recon2}
\end{figure}
\begin{figure}[h]
\begin{center}
    \includegraphics[width=1.0\linewidth]{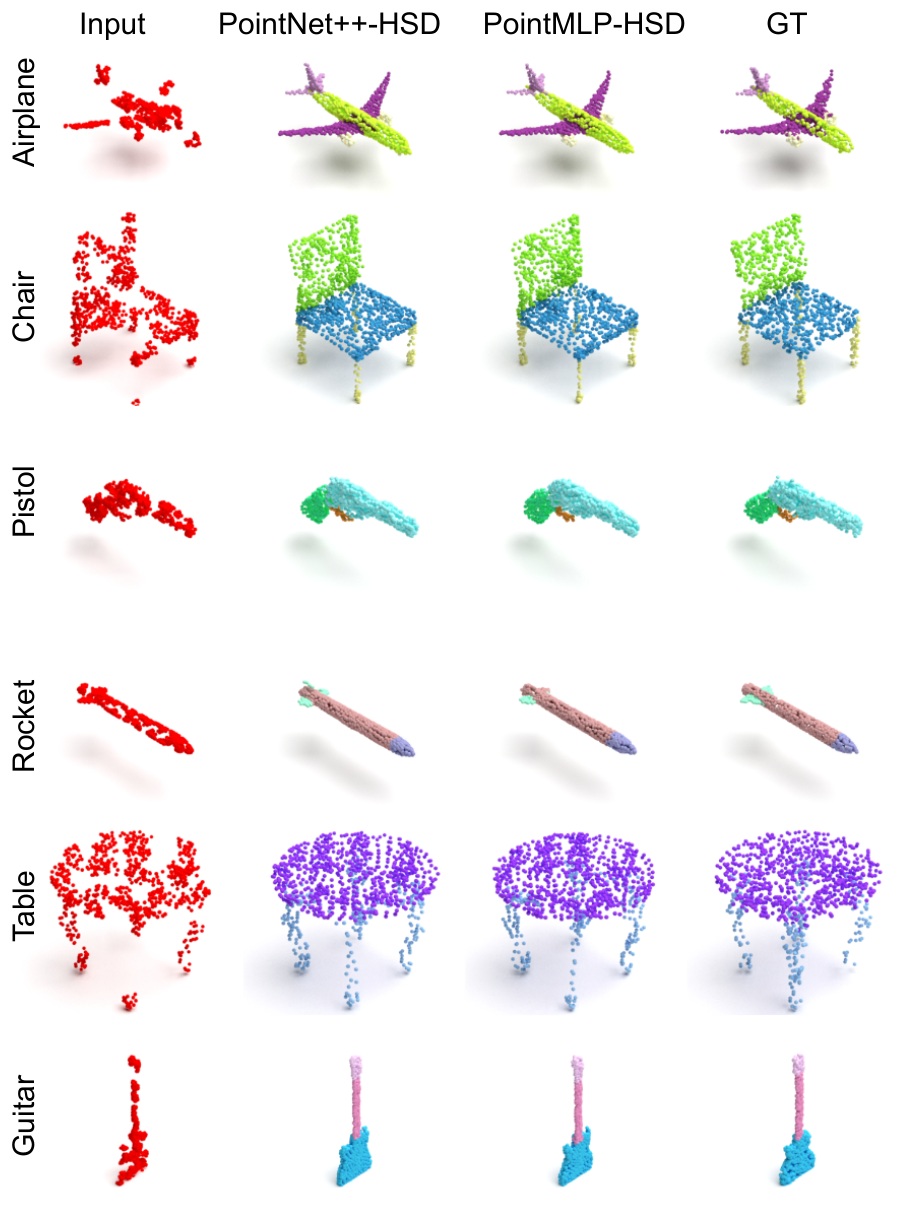}
\end{center}
   \caption{Completed objects of upstream on ShapeNetPart.}
\label{fig:part1}
\end{figure}

Unlike images, probability distributions can only be defined for continuously distributed variables. For point clouds, we assess information by discretizing and evaluating relative information content using KL divergence and cross-entropy. First, we visualize the information plane in Fig.~\ref{fig:quan_ana}(a), where both $I(X;Z)$ and $I(Y;Z)$ steadily increase, eventually reaching a maximum. This is reasonable, since the objective of our MAE is to fully reconstruct $X$ from latent representations $Z$. A similar phenomenon is observed in an autoencoder-based classifier~\cite{lee2021information}, and aligns with other studies, such as \cite{geiger2021information, goldfeld2018estimating}, which indicate that compression is not always required for strong generalization performance. Second, since Fig.~\ref{fig:quan_ana}(a) merely differentiate the information of 3 levels separately, we quantify $I(Z_l;Z_L), l \in [1,2], L=3$ in terms of KL loss in Fig.~\ref{fig:quan_ana}(b). The differences ($\Delta 1$ and $\Delta 2$) between companion levels and the teacher measured by KL divergence ($\mathcal{L}_\mathit{kl}$) decrease progressively. This indicates that information from the deepest level flows to preceding levels, providing them with stronger regularization. The observation that $\Delta 1 < \Delta 2$ further supports the assumption in Sec.~\ref{sec:hsd} that $I(Y;Z_l) \leq I(Y;Z_L)$ for a sparse input. Finally, we utilize cross-entropy to approximate $I(Y;Z_l), l \in [1,2,3]$ in Fig.~\ref{fig:quan_ana}(c), which suggests that the teacher shares the most mutual information to true labels and that information becomes richer as $l$ increases.

Specifically, we use 6 bins to discretize the information $Z_l, l \in [1,2,...,L]$ (i.e., network weights) across the three levels during the initial 100 epochs. With the probabilities of the given data $X$ and labels $Y$, mutual information $I(X;Z_l)$ and $I(Y;Z_l)$ are estimated as follows:
\begin{equation}
\scalemath{0.86}{
\begin{split}
    & I(X;Z_l) = H(Z_l) - H(Z_l|X), \\
    & I(Y;Z_l) = H(Z_l) - H(Z_l|Y).
\end{split}
}
\end{equation}
The procedure for calculating the mutual information and plotting the information plane is included in the source code.

\subsection{Qualitative Visualization}\label{sec:vis}

Figs.~\ref{fig:recon1}, \ref{fig:recon2}, and \ref{fig:part1} visually demonstrate the improvements in reconstructed point cloud quality for ModelNet40, ScanObjectNN, and ShapeNetPart, respectively. For enhanced visualization, Fig.~\ref{fig:recon2} employs the PB\_T50\_RS without background for ScanObjectNN. Notably, some input objects are unrecognizable even for human, highlighting the necessity of the upstream stage for better understanding of 3D objects. Particularly, in Fig.~\ref{fig:recon1}, the reconstruction of the chair’s leg—a crucial feature for accurate classification—is successfully achieved with our HSD methods. In Fig.~\ref{fig:part1}, the reconstructed point clouds exhibit higher quality than the sparsely sampled ground truth.
Since the final objective is object recognition rather than precise completion,  the reconstructed point clouds are restricted to 1,024 points to maintain efficiency in downstream tasks, capturing only the overall shape of an incomplete point cloud.

\subsection{Ablation Study}\label{sec:ab}
\begin{table}
\caption{Ablation studies of regularization on ScanObjectNN (PB\_T50\_RS: object w/o. BG).}\label{tab:ab}
\begin{center}
\begin{tabular}{lcccc}
\hline
\rowcolor{gray!20}
Our $\mathcal{J}$ & Level 1 & Level 2 & Level 3 &  \\
\rowcolor{gray!20}
(PointNet++) & OA/mAcc & OA/mAcc & OA/mAcc & \multirow{-2}{*}{CD} \\
\hline
Baseline & 82.2/80.1 & - & - & 12.11 \\
DSN & 82.4/80.0 & 83.5/81.7 & 84.1/82.8 & 11.51 \\
SCL (HSD) & 82.8/80.5 & 83.6/81.7 & 83.9/83.0 & 11.25 \\
Full (HSD) & 82.8/81.0 & 83.9/82.0 & 84.5/82.7 & 11.17 \\
\hline
\end{tabular}
\end{center}
\end{table}
\begin{table}[h]
\caption{MAE (with PointNet++-HSD as downstream) comparison on SONN w/o BG.}\label{tab:ab2}
\begin{center}
\begin{tabular}{l|ccc}
\hline
\rowcolor{gray!20}
Upstream & OA & mAcc & CD \\
\hline
PMP-Net++~\cite{wen2022pmp}  & - & - & 10.89 \\
\hline
\rowcolor{LightCyan}
Our $\mathcal{J}$ (SnowflakeNet) & 84.5 & 82.7 & 11.16 \\
\rowcolor{LightCyan}
Our $\mathcal{J}$ (PMP-Net++)  & 85.1 & 83.5 & 9.73 \\
\hline
\end{tabular}
\end{center}
\end{table}
\begin{table}[h]
\caption{Evaluation results of the SOTAs (upstream names omitted) and our joint learning ($\mathcal{J}$) methods on ShapeNetPart with $64 \times 16$ scattered points as input.}\label{tab:part}
\begin{center}
\begin{tabular}{l|ccc}
\hline
\rowcolor{gray!20}
Model & mIOU(\%) & cIOU(\%) & CD \\
\hline
PointNet++~\cite{qi2017pointnet2} & 82.9 & 78.8 & \multirow{3}{*}{24.46} \\
PointMLP~\cite{ma2022rethinking} & 83.2 & 79.0 &  \\
PointNeXt~\cite{qian2022pointnext} & \textbf{83.5} & 78.9 &  \\
\hline
\rowcolor{LightCyan}
Our $\mathcal{J}$ (PointNet++-HSD) & 83.3 & \textbf{79.7} & \textbf{24.18} \\
\rowcolor{LightCyan}
Our $\mathcal{J}$ (PointMLP-HSD) & \textbf{83.5} & \textbf{80.2} & \textbf{24.14} \\
\hline
\end{tabular}
\end{center}
\end{table}
\begin{table*}[h]
\caption{Per-class IOU Results of the SOTAs (upstream name omitted) and our joint learning ($\mathcal{J}$) methods on ShapeNetPart with $64 \times 16$ scattered points as input.}\label{tab:part2}
\begin{center}
\scalebox{.86}{
\begin{tabular}{l|cccccccccccccccc}
\hline
\rowcolor{gray!20}
Model & Airplane & Bag & Cap & Car & Chair & Earph & Guitar & Knife & Lamp & Lapt & Motor & Mug & Pistol & Rocket & Skate & Table \\
\hline
PointMLP~\cite{ma2022rethinking} & 79.3 & 78.4 & 78.7 & 72.8 & 87.9 & 67.1 & 89.8 & \textbf{86.4} & 79.9 & \textbf{95.1} & 60.6 & 91.9 & 79.4 & 60.6 & 73.7 & 82.7 \\
PointNeXt~\cite{qian2022pointnext} & \textbf{80.1} & 74.5 & 80.2 & 72.6 & \textbf{88.1} & 72.6 & 89.7 & \textbf{86.4} & \textbf{82.8} & 94.6 & 55.0 & 91.3 & 79.7 & 59.6 & 72.3 & 82.1 \\
\hline
\rowcolor{LightCyan}
Our $\mathcal{J}$ (PointNet++-HSD) & \textbf{80.1} & \textbf{80.6} & 78.8 & 72.3 & 87.8 & \textbf{74.5} & \textbf{91.1} & 85.9 & 79.8 & 95.0 & 58.1 & \textbf{92.8} & \textbf{81.4} & 60.6 & 72.9 & \textbf{83.2} \\
\rowcolor{LightCyan}
Our $\mathcal{J}$ (PointMLP++-HSD) & 79.6 & 77.2 & \textbf{82.1} & \textbf{73.8} & 88.0 & 67.5 & \textbf{91.0} & \textbf{86.4} & 79.3 & 94.5 & \textbf{61.7} & \textbf{92.6} & \textbf{81.1} & \textbf{61.2} & \textbf{77.3} & \textbf{83.1} \\
\hline
\end{tabular}
}
\end{center}
\end{table*}

Sections~\ref{sec:exp-sd} and \ref{sec:exp-ipc} present the comparisons between our baseline models and those based on the HSD mechanism, indicating the superior performance of the proposed approach. The results further demonstrate the scalability of the HSD mechanism in scenarios involving flawed or rapidly scanned point clouds. To validate the effectiveness of HSD, we conduct ablation studies with and without the self-distillation loss, on the PB\_T50\_RS variant (without background) of ScanObjectNN. Here, PointNet++ serves as the backbone for the downstream model, with results presented in Table~\ref{tab:ab}. First, the performances of baseline and DSN (i.e., w/o. HSD as discussed in section~\ref{sec:hsd}) are reported. In addition, we demonstrate the importance of a suitably chosen teacher's code length by employing identical dimensions in the hyperspaces for the last two classification heads, as indicated by the same code length (SCL). Lastly, the full HSD setting provides \textbf{stronger regularization} for the optimization. These two observations suggest that a reasonably longer code length to hold the feature extracted by a larger perceptual field provides more information (i.e., $I(Y;Z_l) < I(Y;Z_L)$), and maximizing the mutual information $I(Z_L;Z_l)$ leads to stronger regularization. Specifically, compared to DSN, i.e., strong regularization, our HSD reduces reconstruction error by 3\% and increases overall accuracy by 0.4\%, confirming the effectiveness of the proposed regularization term.

Worth mentioning, the selection of MAE in our method is also plug-and-play, as shown in Table~\ref{tab:ab2}, where we use PMP-Net++~\cite{wen2022pmp} as the upstream. Note that PMP-Net++ is simplified similarly as SnowflakeNet. With the same downstream network, i.e., PointNet++-HSD, Our $\mathcal{J}$ model with a more powerful upstream consistently improves all three metrics. This further verifies that HSD enables MAE to rearrange high-level details that cannot be reconstructed by the MAE alone.

\subsection{Part Segmentation}
To demonstrate the generality of our framework, we compare our methods with the SOTAs for part segmentation in Table~\ref{tab:part}. Since PointMLP~\cite{ma2022rethinking} and PointNeXt~\cite{qian2022pointnext} use both coordinate and normal information for training, we retrain them with only the coordinates (2,048 points) on the ShapeNetPart training set and test the results generated from the upstream. Both the training and testing sets are normalized for consistency. Under an unfair condition, i.e., with merely half of the input (1,024 points) for training, our simplest model $\mathcal{J}$ (PointNet++-HSD) already outperforms PointMLP. This suggests that our framework possesses the potential to elevate the performance of a model to that of a subsequent generation, and reaffirms the generalization ability of our framework for scattered point clouds on the part segmentation task. We present the per-class segmentation IOUs in Table~\ref{tab:part2}. Specifically, the per-class IOUs are obtained from the checkpoints of the best cIOU for fair comparisons. 

\noindent\textbf{Implementation of Part Segmentation. }
The network structure for part segmentation closely resembles that of classification, with the key distinction that the multi-heads leverage a class-agnostic shape code for self-distillation. However, Calculating loss and evaluating segmentation on incomplete point clouds are non-trivial. Metrics such as per-point accuracy and mean IoU require precise point ordering, as in the ground truth data. Since the generated point cloud lacks this exact order, directly computing these metrics becomes infeasible. The reconstruction loss $\mathcal{L}_\mathit{rec}$ offers a workaround by defining a bidirectional mapping—though not on a per-point basis—between the predicted and ground truth point clouds. This mapping enables point index alignment between the two sets.

\subsection{Failure Case and Future Work}
ModelNet40 contains a diverse set of classes, some of which have similar shapes (e.g., table and bench), which we refer to as the similar-class issue. We consider this a partial failure case, particularly under conditions of low to medium sparsity. Additionally, extremely complex shapes may pose challenges for accurate recovery. 

To address such cases in future work, supplementary prompts (e.g., text) could be incorporated, as seen in BERT-based~\cite{devlin2019bert,wu2024towards} or CLIP-based~\cite{radford2021learning,zhang2022pointclip} methods, which could assist in distinguishing subtle shape variations.

\section{Conclusion}
Recent state-of-the-art models have shown limited  effectiveness on incomplete point clouds. 
In this paper, we have proposed a cascaded network that tailors the upstream and downstream components for analyzing scattered point clouds. In our method, MAE serves as a rudimentary form of object completion, focusing on low-level features, while HSD further refines the understanding of this preliminary reconstruction by emphasizing high-level features, making it particularly well-suited for scattered point cloud analysis. Our experiments demonstrate that our joint learning design generalizes well to scattered point clouds, indicating that the reconstructed objects enhance environmental understanding, particularly in scenarios where perfect scanning is impractical during data collection.
Moreover, our proposed hierarchical self-distillation improves the capacity of hierarchy-based classification models by maximizing the mutual information between layers, thereby further regularizing the optimization process through information flow.

\bibliographystyle{IEEEtran}
\bibliography{jsen.bbl}

\end{document}